\documentclass{article}
\pdfoutput=1

\usepackage{arxiv}

\usepackage[utf8x]{inputenc} % allow utf-8 input
\usepackage[T1]{fontenc}    % use 8-bit T1 fonts
\usepackage{hyperref}       % hyperlinks
\usepackage{url}            % simple URL typesetting
\usepackage{booktabs}       % professional-quality tables
\usepackage{amsfonts}       % blackboard math symbols
\usepackage{nicefrac}       % compact symbols for 1/2, etc.
\usepackage{microtype}      % microtypography
\usepackage{lipsum}		% Can be removed after putting your text content
\usepackage{makeidx}
\usepackage{graphicx}
\usepackage{amsmath,amssymb} % define this before the line numbering.
\usepackage{color}
\usepackage{amsmath,amssymb}
\usepackage{textcomp,marvosym}
\usepackage{cleveref}
\usepackage{float}

\title{Predicting Landscapes from Environmental Conditions Using Generative Networks}

\date{May, 2019}	% Here you can change the date presented in the paper title
\author{
  Christian Requena-Mesa \\
  Climate Informatics, Institute of Data Science \\
  German Aerospace Center (DLR) \\
  Germany, Jena \\
  \texttt{crequ@bgc-jena.mpg.de} \\
   \And
  Markus Reichstein \\
  Max-Planck-Institute for Biogeochemistry \\
  Germany, Jena \\
  \texttt{mreichstein@bgc-jena.mpg.de} \\
   \And
 Miguel Mahecha \\
 Max-Planck-Institute for Biogeochemistry \\
 Germany, Jena \\
 \texttt{mmahecha@bgc-jena.mpg.de} \\
   \And
Basil Kraft \\
Max-Planck-Institute for Biogeochemistry \\
Germany, Jena \\
\texttt{bkraft@bgc-jena.mpg.de} \\
   \And
Joachim Denzler \\
Computer Vision Group \\
Friedrich-Schiller-Universit\" at \\
Germany, Jena \\
\texttt{joachim.denzler@uni-jena.de} \\
}

\begin{document}
\maketitle

\begin{abstract}
	Landscapes are meaningful ecological units that strongly depend on the environmental conditions. Such dependencies between landscapes and the environment have been noted since the beginning of Earth sciences and cast into conceptual models describing the interdependencies of climate, geology, vegetation and geomorphology. Here, we ask whether landscapes, as seen from space, can be statistically predicted from pertinent environmental conditions. To this end we adapted a deep learning generative model in order to establish the relationship between the environmental conditions and the view of landscapes from the Sentinel-2 satellite. We trained a conditional generative adversarial network to generate multispectral imagery given a set of climatic, terrain and anthropogenic predictors. The generated imagery of the landscapes share many characteristics with the real one. Results based on landscape patch metrics, indicative of landscape composition and structure, show that the proposed generative model creates landscapes that are more similar to the targets than the baseline models while overall reflectance and vegetation cover are predicted better. We demonstrate that for many purposes the generated landscapes behave as real with immediate application for global change studies. We envision the application of machine learning as a tool to forecast the effects of climate change on the spatial features of landscapes, while we assess its limitations and breaking points.
\end{abstract}
\begin{center}
	Accepted as a conference paper at GCPR2019 \\
\end{center}

% keywords can be removed

\section{Introduction}

The Earth's land surface can be considered a mosaic of landscapes \cite{forman1995some}. Landscapes are the material-physical entities that comprise the structures of nature \cite{haase1983current}: ecological meaningful units that have a characteristic ordering of elements \cite{mucher2010new}. Landscapes result from the long-term interaction of abiotic, biotic and anthropogenic processes. The relation between landscapes and the climatic, geological, and anthropogenic factors is, however, rather conceptual. The totality of interactions and processes that determine the landscapes are impossible to simulate numerically as of today. This fact holds true to such extent that, to the best of our knowledge, landscape imagery prediction is yet to be attempted. We aim to analyze whether the relation between forming factors and landscapes can be mapped with a statistical method. Our goal is to reconstruct the 2D aerial view (multispectral) of the landscapes from a set of 2D environmental conditions. Furthermore, we assess the use of predicting landscapes as a tool for climate change and landscape change studies.

The study of Earth at the landscape scale gained momentum in the last decades benefiting from the use of geographic information systems and the high availability of remotely sensed imagery \cite{franklin2000quantification,groom2006remote,kerr2003space,newton2009remote,simmons1992satellite}. Remotely sensed images are a measure of the radiation reflected by the surface. The observed reflections at certain wavelength are information rich snapshots that can be used to diagnose features such as land-cover type, ecosystem spatial structure, vegetation health, water availability or human impact \cite{brando2003satellite,getzin2012assessing,otterman1977anthropogenic}. Predicting the aerial image comes close to predicting the landscapes and their spatial arrangement. From the satellite image, one could derive many high level aspects of the landscapes and ecosystems with existing earth observation tools.

Landscapes are formed by a wide range of components and processes. Factors that determine a landscape can be categorized into largely independent ones (e.g. climate or geology) and dependent ones (e.g. soil or vegetation). A change on the independent factor leads to a change of the dependent ones, for example, changes on abiotic factors generally lead to changes in biotic components (such as shifting position or composition). Previous work to classify the landscapes has determined and ranked the forming factors by importance \cite{bunce1996land,klijn1995hierarchical,mucher2010new}

\begin{equation}
\label{eq:components}
L = f(C,G,H,S,V,F,U,S) .
\end{equation}

% too complicated - just state what model we use, which predictors etc
Where $L$ is the Landscape, $C$ is climate, $G$ the geology and geomorphology, $H$ the hydrology, $S$ the soil, $V$ the vegetation, $F$ the fauna, $U$ the land use and $S$ the landscape structure. Developing over the work of \cite{mucher2010new} we can further reduce the conceptual relationship into the essential independent forming factors 
\begin{equation}
\label{eq:formation}
L = f(Clim,Geo,AI) .
\end{equation}

Where $Clim$ is the broadened climate, $Geo$ is the lithology and topography and $AI$ are the anthropogenic interventions. As the dependent factors (e.g. soil or vegetation) can be thought of as a function of the independent ones (climate and geology) direct knowledge is not strictly necessary. In addition we broad the definition of climate to encompass all of the meteorological hydrology.

Mechanistic or statistical approaches are scarce, or only address certain aspects (e.g. geomorphological models). Advances in deep learning allow for unsupervised content-based data driven modeling, i.e, neural networks capable of learning the relationship between the spatial features present in the input and output from available data \cite{isola2017image}. Ideally, these networks can accommodate the non-linearities that best approximate the functional relation between environmental factors and landscapes generating realistic spatial representations of the landscapes. We aim to demonstrate that it is possible to predict landscapes -as seen from space- that behave as real for hypothetical environmental conditions. We attempt to map the climatic, topographic and anthropogenic factors onto sentinel-2 visible and near-IR bands using a conditional generative adversarial network (cGAN)  \cite{mirza2014conditional}. We will assess its limitations and usability as a tool for climate change studies. One of the main applications envisioned for the proposed approach is forecasting landscape change under future climate projections.

\section{Materials and Methods}
% For figure citations, please use "Fig" instead of "Figure".
Generative adversarial networks (GANs) estimate a generative model via an adversarial process in which two neural networks are trained simultaneously: a generative network $G$ that captures the data distribution and a discriminative network $D$ that estimates the probability of a sample coming from the training set rather than from $G$. Both networks are co-trained: the network $G$ tries to maximize the probability of $D$ making a mistake while the network $D$ tries to discriminate data generated by $G$ from true samples \cite{Goodfellow2014}.

In addition, cGANs learn a mapping from input conditions and $r$ probabilistic latent space to the output. Later developed topologies such as the U-Net GAN \cite{isola2017image} allow for the input conditions to have two spatial dimensions. With the use of skip-connection between symmetrical convolutional and deconvolutional layers, the conditions do not only determine the features that shall be present in the output, but where those must appear. This is important since, the location of a feature on one of the conditioning variables, e.g., a mountain range in the altitude predictor, must be reflected on the same location on the generated landscape. The mapping from the probabilistic space is relevant since the conditioning variables do not deterministically determine landscapes.

\subsection{Problem and notations}

%We represent the images in the form of ordered pixels $𝑆 = (v_{ij})_b$ in a space of two spatial dimensions $(ij)$ and $b$ spectral bands. In the same manner we represent the environmental conditions as a set of ordered pixels $C = (v_{ij})_p$ with $p$ variables representing $Clim,Geo and AI$ and the same spatial dimensions as the imagery. 
To model the problem’s uncertainty, we define the ground truth as a probability distribution over the imagery conditioned on the set of environmental conditions $C$. In training we have access to one sample of the target landscapes for each set of environmental conditions. We train a neural network $G$ to approximate the sought function \cref{eq:formation}, returning landscapes as seen from space when fed with the environmental conditions

%Fix the stuff  respect the r and the learning of the multiple images: it does not totally work with the U-Net gan used: but it might with other newer architecures.
%\begin{equation}
%\label{eq:condsampler}
%𝑃( \cdot|𝐶): 𝑆 = 𝐺(𝐶, 𝑟; \theta) .
%\end{equation}
\begin{equation}\label{eq:approx}
𝐺(𝐶, 𝑟; \theta) \approx f(Clim,Geo,AI) .
\end{equation}

where $\theta$ denotes the network parameter and $r$ is a random variable from which to map the multiple plausible outputs. During test time multiple samples of $r$ could be used to generate different plausible landscapes for the same set of environmental conditions.

\subsection{Data}

%[[[[Disclaimer? The selection of the predictor datasets for experimental tests is an arbitraty decision. In general, as long as $Clim$, $Geo$ and $AI$ are well represented any data source should output similar results (care of resolution, etc).]]]]

%Good! Describe what has been done
Satellite imagery sensed on April 2017 by Sentinel-2 was matched with 32 environmental predictors representing the $Clim$, $Geo$, and $AI$ forming factors for $94,289$ locations. The dataset covers $~10\%$ of the emerged Earth surface on 1857 blocks of $110×110$ Km randomly distributed across the planet. Each location will serve as a single sample with $256×256$ pixels, 32 input environmental variables and 4 output multispectral variables. %Given text limit constraints we give a more detailed description in the supplementary material. The dataset is made publicly available.
Climatic variables ($Clim$) were represented by a subset of WorldClim v2 \cite{hijmans2015worldclim}. Altitude and lithology ($Geo$) variables, were represented by STRM v4 \cite{jarvis2008hole} and GLiM \cite{hartmann2012new}. In addition, we used three of the GlobeLand30’s \cite{jun2014china} classes as a proxy for anthropogenic large scale interventions ($AI$). All of the environmental variables were resampled to $256\cdot256$ pixeles to match the resolution of the imagery.
%Nonetheless, for a project of this size it is practially imposible to get the perfect dataset. In adition, we believe one of the requisite of the model should be to deal with such difficulties and learn the significant relations.

%%%%%%%%%%%%%%***Release the dataset: 166 Gb makes it difficult!***\\
%%%%%%%%%%%%%%***It is set and ready on the work2/SentMegadata folder, still not sure how the releasing can be done***

%[[LOTS OF PREPROCESSING IN USALLY HOWEVER...The high collinearity of climatic inputs was not dealt with, we use a complex deep learning model in order to avoid preprocessing biases. The network has to learn what features to extract from the inputs. In the same manner, there is an slight overrepresentation of deserts in the dataset:one of our assumptions is that the network will learn how to best assimilate the data.]]

\subsection{Experimental Design}

Two experiments were designed to asses 1) the ability of the proposed approach to generate landscape imagery and 2) its generalization capability under different spatial block designs to find limitations and describe possible consequences of extrapolation.

\textbf{Experiment One} We first compare five models of different architectural complexity, from a fully connected network to the complex cGAN detailed in the models subsection. For this experiment train and test samples were drawn from a completely randomized pool, 80\% of which were used for training and 20\% for testing. %Under complete randomization the test dataset is expected to be well represented by the train data, therefore we are comparing the different models on an ideal scenario. 
With this experiment we expect to unveil whether there is a significant performance improvement between the proposed cGAN approach and baseline machine learning methods.

\textbf{Experiment Two} We trained the cGANs of different complexity and the baseline fully connected model under %different . Landscapes that are close on the surface of the planet tend to be similar; therefore, training with adjacent landscapes to the ones to be predicted -as in the first experiment- leaks part of the test set. In this case we compare the performance of the models under 
different spatial block designs. Landscapes that are close on the surface of the planet tend to be similar \cite{roberts2017cross}. We split the train and test sets into 3 experimental treatments: a complete randomization of the test and train sets as a baseline; a block design where the test locations must be at least $100$ km from the closest training sample; and a third treatment where landscapes in the Americas are predicted by a model trained solely with Asia, Africa and Europe. We expect to be able to measure the negative consequences of overfitting that might arise when extrapolating to locations far from those used for training, and thus, combinations of input conditions never seen during training.

\subsubsection{Models}\label{sec:models}
Two cGANs were trained inheriting the architecture of the original pix2pix network \cite{isola2017image}: a low complexity one (GAN 1 Gb) with few learnable filters per layer and a high complexity one (GAN 7 Gb), named after the total size of the weights on disk. Alongside with the cGANs a fully connected model lacking spatial context was trained as a baseline. In addition, two handicapped cGAN models were trained in order to compare equally complex models that lack one of the key features. One handicapped cGAN was trained over a modified train set with no spatial features due to random permutation of pixels. It is to expect that this handicapped cGAN will not take advantage of convolutions. The second handicapped cGAN was deprived of the discriminator loss. We would expect it to fail to produce landscapes with sharp photointerpretable features. Further detail of the networks' architectures can be found in the supplementary material.

\subsubsection{Analysis}
%or \subsection*{Analysis}

%Consider rewording: short this human thingy intro.
Domain experts can visually determine whether a pair of satellite images resemble the same ecosystem, have a similar climate or have a relatable landscape structures. Experts can identify if a satellite image is realistic or faulty. However, human perception based metrics are costly, time inefficient and prone to bias; therefore, computerized metrics are needed for the objective comparison of models of results over thousands of samples. The generated landscapes must resemble the target ones, however the generated landscapes do not have to match the targets pixel to pixel. Per pixel error metrics are not adequate since the features of interest on landscapes are of supra-pixel scale, and can appear on different places in the generated imagery.

We tested the generated landscapes by comparing the high level landscape patch metrics \cite{cardille2017understanding} to the target landscapes. Landscape metrics are typically used for the objective description of landscape structure. These are central to the study of landscape ecology and biodiversity and habitat analysis  \cite{uuemaa2009landscape}. We make use of landscape level patch metrics as a mean to compare landscape composition and structure. While our quantitative analysis is focused on the landscape metrics, it is important to note that predicting these metrics is not the objective of our work. Our objective is to generate landscape images that behave realistic, and we make use of the landscape metrics as a mean of automatizing the evaluation.

In order to compute patch metrics of a landscape, segmentating into landscape units is needed first. These landscape units typically have a semantic meaning such as \textit{forest} or \textit{industrial} and the segmentation process is often carried out by humans. However, in an effort to automatize the analysis, we opted for an unsupervised non-semantic segmentation. The generated and the target satellite imagery was unsupervisedly segmented via K-means clustering using the red, green, blue and near-infrared bands as input variables. The individual pixels were segmented into 20 clusters. A value of $K=20$ was selected since this is a typical number of land cover units seen in semantically segmented land cover products \cite{fox2005land}. For robustness, the analysis was repeated under different landscape segmentations, clustering it in an undersegmentation scenario (K=8) and oversegmentation scenario ($K=60$).A total of 8  K-means were trained using $3,000,000$ pixels ($3\%$ of total) extracted randomly from $1,500$ randomly selected target images ($~8\%$ of total). Once trained, in a cross-validation fashion, segmentation was performed for $8$ randomly selected subsets of size $1,500$ sample pairs (each pair consisting of a target and generated landscape). 

As landscape metrics are often redundant \cite{cushman2008parsimony}, we selected five representative landscape level metrics of diverse nature based on expert criteria: Shannon diversity index, patch cohesion, connectance, mean fractal dimensionality and effective mesh size. Landscape metrics were computed using FRAGSTATS v4 \cite{mcgarigal2012fragstats}. %Normalized Difference Vegetation Index (NDVI) was computed for each pair using red and infrared bands. 
Our final evaluation measure is the robust biweight midcorrelation \cite{wilcox1992robust} computed as in \cite{langfelder2012fast} between generated and target landscapes for patch metrics.% and NDVI.

%Though the use of landscape metrics is questioned nowadays, comprehensivensive meta-analysis \cite{uuemaa2009landscape} shows correlation between landscape metrics and biological variables indicating biodiversity and habitat structure, or others such as water quality or landscape aesthetics. 
%Two experiments. We want to test how do GANs compare to less complex models by using handicapped versions with the same complexity and also to a shallow learning fully connected model (pixel based), serving as a base line or reference of what we could expect using these pixel based approaches (RF, SVM, etc).]]]

% Place tables after the first paragraph in which they are cited.

\section{Results}
\subsection{Quantitative Analysis}
\subsubsection{Experiment One}
\begin{figure}[!h]
	\centering
	\includegraphics[scale=0.7]{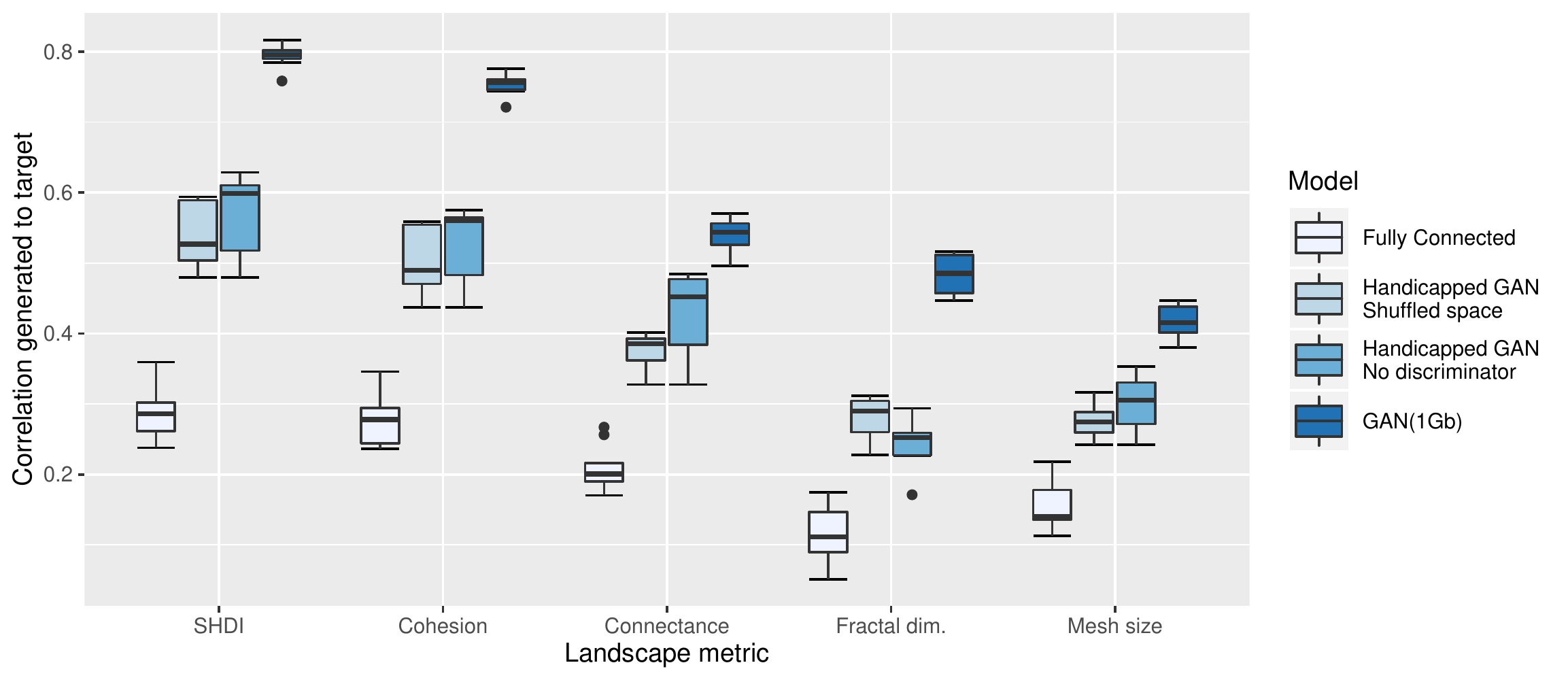} %Do not include figures on the submission TeX and PDF
	\caption{{\bf Intermodel comparison (20 landcover units).}
		Correlation between landscape metrics of generated and target landscapes. Shannon diversity index (SHDI), patch cohesion, patch connectance, average patch fractal dimensionality and effective mesh size were computed for both, real landscapes, and landscapes generated given the environmental conditions on the test locations.}
	\label{models_metrics_comparison}
\end{figure}

%%Landscape metrics
%%Experiment 1
%%On our first experiment we compare our approach and the baselines's landscape metrics on a test set well represented on the train set. 
Landscapes level metrics' correlation between generated and target landscapes on our first experiment can be seen in \cref{models_metrics_comparison}. The proposed model (GAN 1 Gb) is best at producing landscapes whose patch level metrics resemble the target landscapes.
The handicapped models, in spite of having the same number of total weights as the proposed GAN, fail to reproduce the landscape metrics to the same degree. This indicates both, the use of a discriminative training and the capability of mapping spatial features, are key for a high performace. The pixel based `FC' model is the least capable of generating landscapes whose landscapes composition and structure resemble the targets. Results evidence that for best performance, models that can make explicit use of spatial neighboring  on the input features and can generate the landscape as a whole rather than per pixel are needed. In addition, the best performing models are those using a discriminative loss, rather than simple per pixel error metrics. Further evaluation with undersegmented and oversegmented landscapes also lead to similar results (supplementary figures 2 and 3). This indicates that the findings are not dependent on the segmentation process used for calculating the landscape metrics.

\begin{figure}[!h]
	\centering
	\includegraphics[scale=0.8]{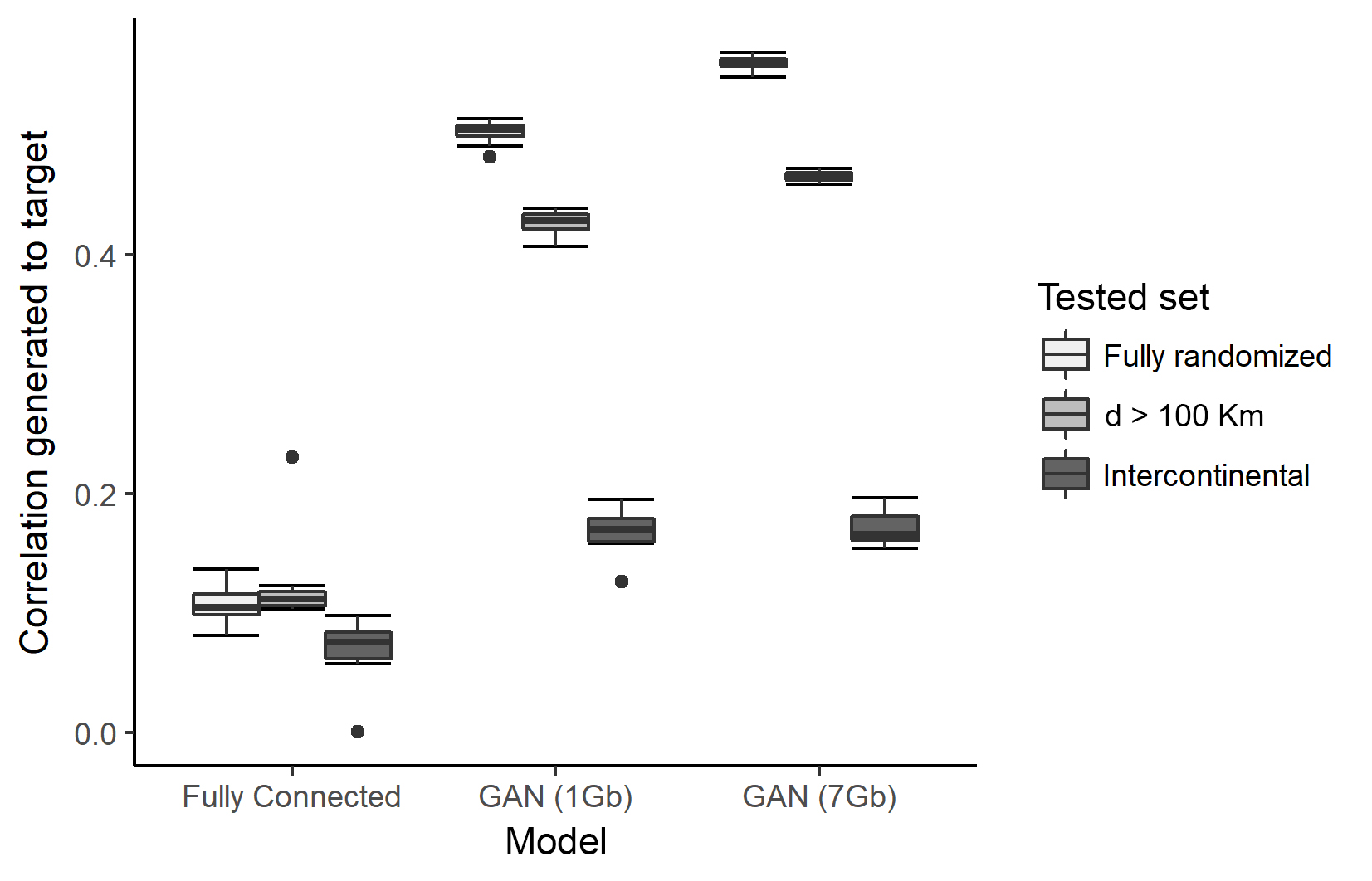} %Do not include figures on the submission TeX and PDF
	\caption{{\bf Model generalizability.}
		Correlation between target and generated landscapes' averaged patch metrics. The baseline FC and the two models representing the proposed approach (GAN 1 Gb and GAN 7 Gb) are compared over 3 different test sets. The fully randomized test set is well represented by the training data. The intercontinental set has sample locations on the Americas for models trained with locations on Asia, Europe and Africa. In the intermediate case, the locations are at least 100 Km apart from the closest location used for training.}
	\label{fig:complexity_metrics_comparison}
\end{figure}

\subsubsection{Experiment Two}
On our second experiment we compared our approach and the baseline on a test set that is vaguely represented by the test set due to distance between samples, some extrapolation occurs at prediction time. Landscapes level metrics' correlation generated to target for the proposed model and the baseline can be seen in \cref{fig:complexity_metrics_comparison}. The proposed GAN models (1 Gb and 7 Gb) outperform the simple baseline method FC. We observe all models suffer some kind of performance decay as location of the testing set samples is further from the train locations. The FC model's performance decays the least with distance, this is to expect of a model that does not suffer from overfitting. The performance of the more complex GAN models decays strongly with extrapolation. While the GAN 7 Gb model slightly outperforms the GAN 1 Gb model on the fully randomized test set and the short distance test, performance becomes similar on the harder intercontinental test. This indicates some level of overfitting to train locations. While the performance decay due to extrapolation is the largest on the GAN 7 Gb, it is to note that it still outperforms the simpler method that does not use spatial context.

\subsubsection{Normalized Difference Vegetation Index Prediction}
\begin{table}
	\centering
	\caption{
		{\bf Generated to target correlation for the Normalized Difference Vegetation Index}}
	\begin{tabular}{llllll}
		& FC   & Spatially shuffled    & No discriminator  & GAN (1 Gb)          & GAN (7 Gb)         \\ \hline
		\multicolumn{1}{l|}{Fully randomized}       & \multicolumn{1}{l|}{0.934} & \multicolumn{1}{l|}{0.989} & \multicolumn{1}{l|}{0.995} & \multicolumn{1}{l|}{0.991} & \multicolumn{1}{l|}{0.965} \\ \hline
		\multicolumn{1}{l|}{d $>$ 100 Km} & \multicolumn{1}{l|}{0.903} & \multicolumn{1}{l|}{}      & \multicolumn{1}{l|}{}      & \multicolumn{1}{l|}{0.980}  & \multicolumn{1}{l|}{0.978} \\ \hline
		\multicolumn{1}{l|}{Intercontinental} & \multicolumn{1}{l|}{0.716} & \multicolumn{1}{l|}{}      & \multicolumn{1}{l|}{}      & \multicolumn{1}{l|}{0.749}  & \multicolumn{1}{l|}{0.718} \\ \hline
	\end{tabular}
	\label{NDVI_table}
\end{table}

Summary statistics for the prediction of the overall amount of vegetation for all models can be seen in \cref{NDVI_table}. As NDVI is a simple ratio between different spectral bands, these results can also be understood as the ability of the models to predict overall reflectance.

The simpler models perform close to the GAN models. This effect might be due to the unnecessity of context and spatial features as NDVI is averaged across the image. While by a small margin, the more complex models (GAN 1 Gb and GAN 7 Gb) outperform the FC baseline models. The higher performance cannot be attributed to the capability of these models to learn spatial features or the discriminative loss, since both handicapped models also outperform the baseline. Therefore, the most plausible cause for the higher performance is the sheer size of the models. Similar to what has been observed for the landscape metrics, the simpler baseline method performance does not decrease as steeply as the GANs when predicting intercontinentally; however, the absolute performance of the GANs is still superior even when extrapolation is done.

\subsection{Visual Analysis}
\begin{figure}[!h]
	\centering
	\includegraphics[scale=1.4]{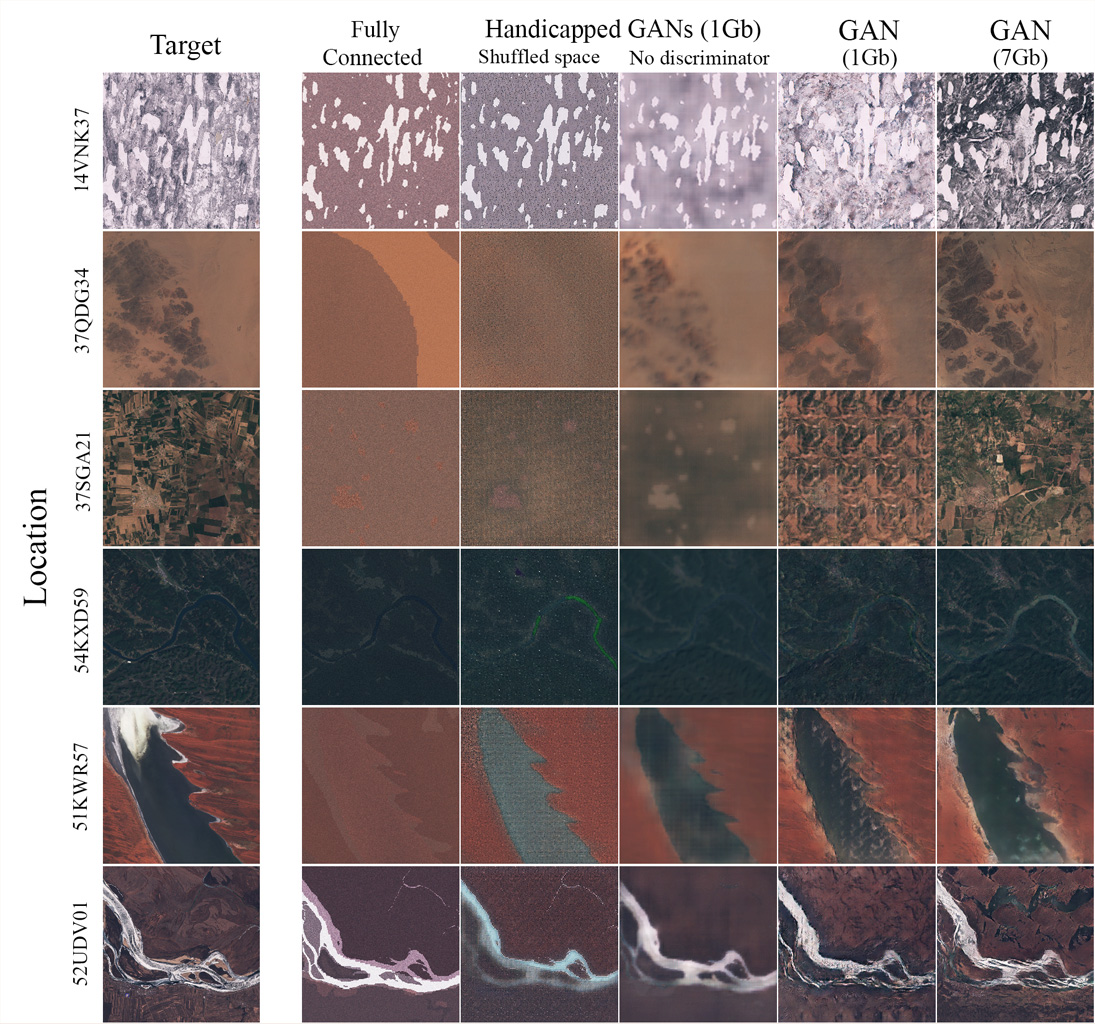} %Do not include figures on the submission TeX and PDF
	\caption{{\bf Sample generated landscapes across the tested models}
		Visible bands on test locations belonging to the fully randomized test set. Model complexity increases from left to right. The six displayed locations were drafted at random.}
	\label{fig:intermodel_imagery_comparison}
\end{figure}
%%EXPERIMENT 1
\subsubsection{Experiment One} 

We first present generated and target landscapes on an ideal scenario where the train dataset represents very well the cases in the test set due to spatial proximity (\cref{fig:intermodel_imagery_comparison}). Landscape reconstructions based on per pixel mapping (FC model) gives overly a close rendition of reflectance across all bands, i.e., colors resemble the target with few exceptions (as seen in sample 14VNK37). Nonetheless, these generated samples lack the characteristic spatial features of landscapes. This is visible on an agricultural area where the model produces an image that lacks the spatial features of agricultural fields (sample 37SGA21). These samples can hardly be understood as satellite imagery, preventing photointerpretation. Moreover the spatial heterogeneity seen in the output of the fully connected model is directly determined by the spatial heterogeneity of the input condition variables.

The spatially shuffled handicapped GAN faces similar visual problems as the FC model. Although overall colors seem accurate, it still fails to project the expected spatial features of real landscapes as it lacks content-based generation. The outputs are noisier than the fully connected model, possibly due to the spatially shuffling of pixels on the training set and the discriminative loss, i.e., noisy samples, similar to those used for training, had a smaller discriminative loss. The 'No discriminator' handicapped GAN, in contrast, can make use of convolutions but lacks the discriminative training. It generates outputs that contain spatial features loosely resembling those of real landscapes; however, these are smoother. This is an expected artifact when using solely mean squared error as training loss.

GAN models (1 Gb and 7 Gb) generate crisper images. These contain spatial features that humans can recognize as part of landscapes that are not determined by the spatial features in the predictors. The lower complexity model (GAN 1 GB) seems to lack the ability to learn enough spatial features on its generator, and under some circumstances generates mosaicking patterns (as seen in sample 37SGA21) while the model with higher number of convolutional filters per layer (GAN 7 Gb) seems to be able to cope better with spatially homogenous predictors. However, it is still susceptible to generate visually faulty landscapes as seen in sample 52UDV01.
Under simple visual inspection, the proposed models (GAN 1 Gb and 7 Gb) seem to be the ones generating imagery that is most readily accessible to photointerpretation.

%%EXPERIMENT 2
\begin{figure}[!h]
	\includegraphics[scale=1.55]{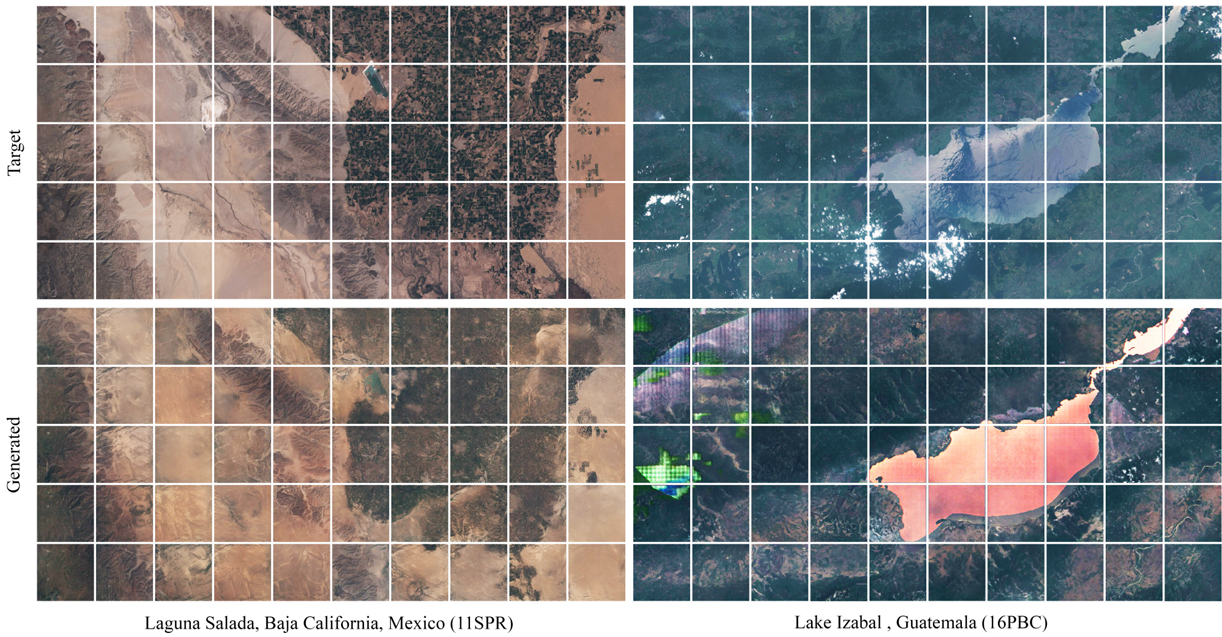} %Do not include figures on the submission TeX and PDF
	\caption{{\bf Generated test samples by a GAN (7 Gb) for the intercontinental test set.}
		The model used to generate the landscapes was trained solely with imagery of Europe, Africa and Asia, however the input conditions given to generate the images occur in the Americas. Each square image was generated independently and mosaicked to display a larger area.}
	\label{fig:Mounted_mosaic}
\end{figure}

\subsubsection{Experiment Two} We trained the models with samples from Europe, Africa and Asia to later generate samples for environmental conditions occurring on the Americas. Samples from a good and a bad case were re-mosaicked for display on \cref{fig:Mounted_mosaic}. While Laguna Salada landscape does not show large scale artifacts, Lake Izabal landscape displays colors and features that are not possible on a real landscape, such as a red water lake or bright green vegetation, as well as, repetitive patterns. %next sentences are discussion

\section{Discussion}
% divide the discussion into method and interpretation discussion
We tackled the prediction of landscapes as seen from space by linking reflectance and environmental conditions with a generative neural network. The proposed approach is able to generate photointerpretable satellite imagery. This is the first time it has been achieved. We unveiled that both, a discriminative loss and spatial context are key for the good performance of the model. In addition, spatial extrapolation to new areas is possible to some extent.

Nonetheless, there are artifacts of different nature in the generated imagery, especially for the test locations that are far from the train locations. We hypothesize these artifacts could be caused by a combination of the following reasons: 1) the deep dreamy appearance might be a negative consequence of the discriminative loss; 2) the model might be overfitting our training set; 3) the input environmental conditions were never seen during training; 4) the processes involved in creating landscapes are not identical in different parts of the planet. These results highlight the importance, when using this approach as a tool, of training the model with a set of samples that is as relevant as possible to the problem to be solved. 

\subsection{Predicting Landscapes as a Tool}

Predicting landscapes could become an important tool for the study of ecological change at the landscape scale. However, further effort is needed as the proposed approach is by no means exempt of flaws and projecting future climate scenarios is only valid under rigorous assumptions.

\begin{figure}[!h]
	\centering
	\includegraphics[scale=0.3]{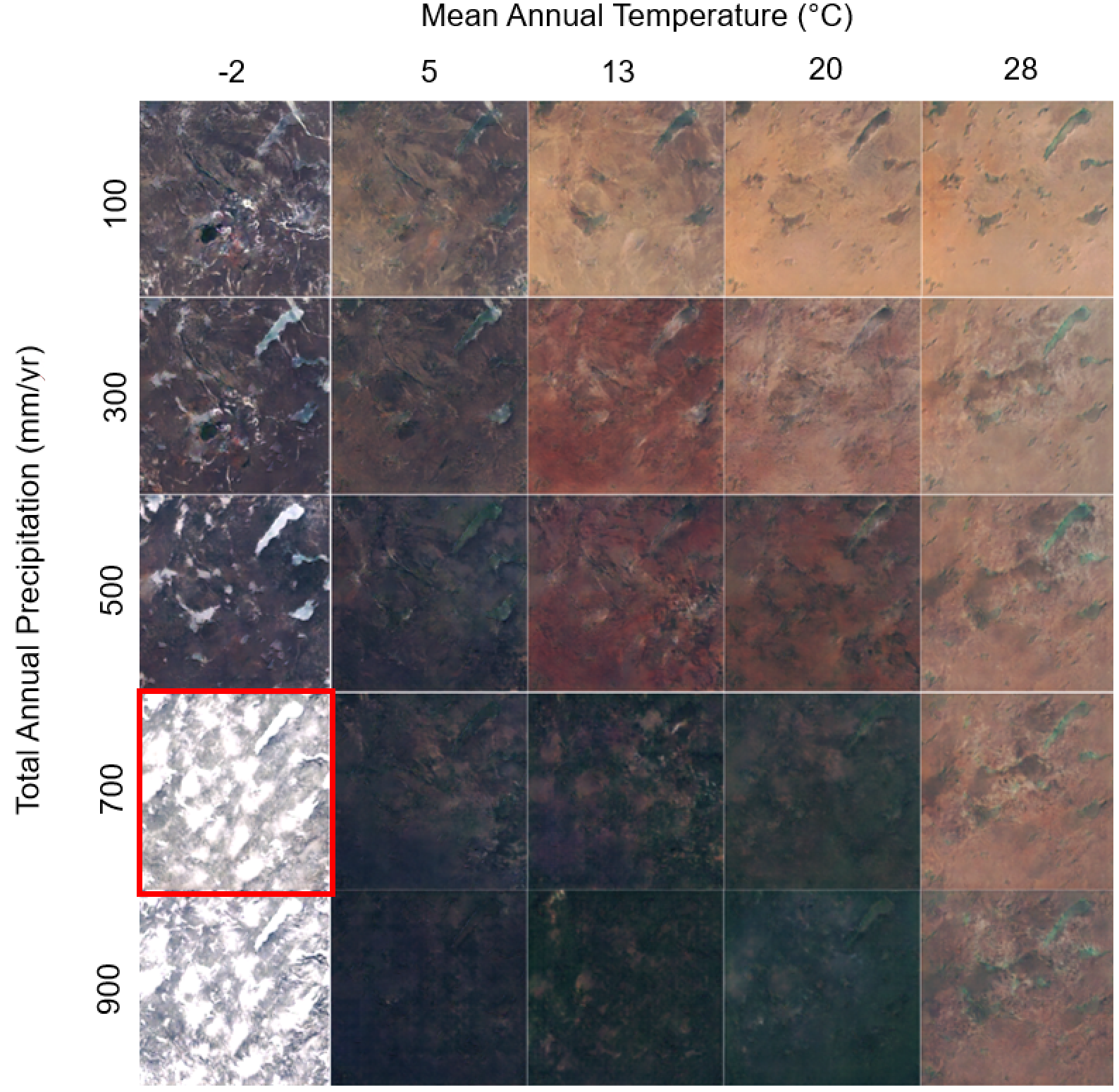} %Do not include figures on the submission TeX
	\caption{{\bf Satellite imagery generated for hypothetical environmental conditions.}
		Annual precipitation and temperature were modified over a real location in northern Canada (56°48'37"N, 106°32'36") in order to display the ability of the model to create novel landscapes for hypothetical climate scenarios. The real landscape is marked in red.}
	\label{fig:generating_landscapes}
\end{figure}

One of the main applications envisioned for the proposed approach is forecasting landscape change under future climate projections. In \cref{fig:generating_landscapes} a sample test location is altered both, annual precipitation and mean annual temperature, in order to display the capability of the model to reimagine landscapes under different environments. Validating the landscapes depicted in \cref{fig:generating_landscapes} is not possible; however, the increase of vegetation cover and the desertification as temperature and precipitation varies is consistent with our expectation. While, made up predictors generate landscapes never seen and thus, difficult to validate, we have tested against predictors found in real world mid distance (more than 100 Km away) and long distance (intercontinentally) and the landscapes generated do resemble the real landscapes to a high extent.

There are, however, limitations to be taken into account before using these models for projecting future landscapes. Our approach, in its current status, does not account for the temporal processes; instead, if we were to predict future scenarios we would be assuming space for time, e.g., sometime warmer in the future must look like somewhere warmer today. However, this assumption only holds if you allow ample time for the landscape to achieve a new steady state after the intervention. When climatic conditions change rapidly, landscape change lags behind, and thus, modelling the landscapes inertia and temporal dynamics is necessary. Also with global change other factors such as $CO_2$ change, having an important effect on vegetation that a static model cannot predict. In addition, in order to generate landscapes for hypothetical environmental conditions, we assume that the system is currently in steady state. However, we know that current climate is trending \cite{allen2014ipcc}. Therefore, since our train set landscapes are not in equilibrium with the environmental conditions, our ability to predict landscapes for the future climates is further detrimented. The data driven modelling of the evolution of landscapes over time given the environmental conditions might server as a fix to most of the flaws previously mentioned.

Although landscapes are shaped by past environmental conditions, for simplicity, we have assumed that present day conditions are sufficient. As a matter of fact, present day conditions may serve as a proxy for past climatic conditions. While this is true for orbital processes, it is not for the tectonic processes or glaciations \cite{zachos2001trends}.

%In cases such as the intercontinental prediction, it is to note that using a pre-trained model and adding a few target samples of the new continent for re-tuning has potential to improve performance while requiring very little new data.

Predicting the landscape patch composition and structure is not the only use that can be made of the imagery. The generated satellite imagery can be potentially used across many disciplines. There are many available tools for different scientific purposes that make use of satellite imagery to gather information. It is unknown whether the quality of the generated imagery is sufficient for these tools to work seamlessly and will require specific testing for the different use cases.

\subsection{Problems}
Predicting landscapes as seen from space is an ill-posed problem. Landscapes are stochastic and chaotic systems and their evolution is conditioned on an unknown initial state. For added difficulty, landscapes have become dependent on complex societal and economical systems that are hard to predict as well. We cannot expect our method or any future method to predict landscapes without flaws and compromises.

We have experienced technical problems. We observe mode collapse. The GAN network we use \cite{isola2017image} does not explicitly model the uncertainty since no $r$ probabilistic space is used during the generative process. Instead, it does so in a weak manner by having an active dropout during test time as explained in \cite{isola2017image}. When fed the same set of environmental conditions, the network outputs nearly identical landscapes. Later developed network architectures, such as those using an implicit probabilistic space and making use of Wasserstein distance during training might mitigate the observed mode collapse. Predicting the full set of landscapes that are plausible for a set of environmental conditions is key in order to have a measure of uncertainty over our prediction.

\subsection{Future Work}
Predicting imagery of landscapes is a valid line of research; furthermore, future improvements in deep learning must be expected making the prediction of satellite imagery more feasible.

\begin{itemize}
	\item Modeling explicitly landscape evolution over time could greatly benefit the usability as a tool. 
	\item Improve landscape comparison metrics: comparing generated and target landscapes is not trivial. Learned perceptual similarity based metrics \cite{zhang2018unreasonable} might be a faster and better option than landscape metrics.
	\item Extract knowledge from the network. Exploring the latent space might give us new insight to the importance of the forming factors, the relation between them and help clustering the surface of Earth landscape-wise.
	\item Build a working stochastic generator. Generate the multiple plausible landscapes that can be arise from each set of environmental factors.
\end{itemize}

\section{Summary and Conclusion}
We have tackled the prediction of landscapes as seen from space by approximating the conceptual model with a generative neural network. The proposed approach demonstrates that a minimum set of environmental conditions is enough to predict landscapes. Our trained model allows to generate close to realistic landscapes for hypothetical environmental scenarios that have some degree of photointerpretability. To the best of our knowledge, this is the first time that environmental predictors are used to infer the aerial view of landscapes.

The predicted images of the landscapes have spatial features that are not dictated by the predictors, but introduced by the generative model. These spatial features add for the interpretability as evidenced by our experiments (\cref{fig:intermodel_imagery_comparison}), making the generated landscapes behave closer to the real ones as evidenced by patch level landscape metrics \cref{models_metrics_comparison}). We contribute our dataset covering 10\% of the emerged surface of Earth and matched with the pertinent 32 environmental predictors (detailed in Supplementary Material) for further development. We believe this is an important step for the data-driven modeling and forecasting of Earth surface.

The use of a discriminative loss and spatial context is crucial in order to generate landscape images susceptible to photointerpretation. We see how a minimum set of with only present day environmental conditions provide enough information to infer the aerial view of a landscape. We demonstrate for the first time that landscapes, as seen from space, can be predicted by pertinent environmental conditions, opening a new data-driven way to study the landscape evolution.

\bibliographystyle{unsrt}  
\bibliography{landscapes}  %%% Remove comment to use the external .bib file (using bibtex).

\section{Supplementary Material}
\subsection{Dataset}

%[[[[Disclaimer? The selection of the predictor datasets for experimental tests is an arbitraty decision. In general, as long as $Clim$, $Geo$ and $AI$ are well represented any data source should output similar results (care of resolution, etc).]]]]

%Good! Describe what has been done
The sample satellite imagery was sensed by the Sentinel-2 and dated in April 2017 for random locations across the globe. We collected $1,857$ level-1C tiles of $110×110$ km$^2$ each from latitude 56°S to 60°N with less than $20\%$ cloud cover. Each tile was subdivided into $10×10$ cells, each cell is treated as a single sample of $11×11$ km. The samples missing data total or partially were discarded. A total of $94,289$ samples were obtained and resampled to $256\cdot256$ pixels, resulting in 43 m/pix resolution. Only blue, green, red and near-infrared spectral bands were considered.

% Samples from Sentinel 2. Each 1C level intake was subdivided into a 10x10 grid. Naming of the files correspond to the sentinel grid adding a 2 digits at the end for the grid position.
Environmental predictors were collected and matched to each sample location. Climatic variables ($Clim$) were represented by a subset of WorldClim v2 \cite{hijmans2015worldclim}: annual precipitation, mean annual temperature, precipitation of the wettest month, precipitation seasonality, precipitation of the driest month, maximum annual temperature, minimum annual temperature, mean diurnal range, isothermality, temperature seasonality and annual temperature range. Altitude and lithology ($Geo$) variables, were represented by STRM v4 \cite{jarvis2008hole} and GLiM \cite{hartmann2012new} of which only the 18 high level lithological classes were used as predictors. In addition, we used three of the GlobeLand30’s \cite{jun2014china} classes: agriculture, artificial and urban, and waterbody, as a proxy for anthropogenic large scale interventions ($AI$). All of the environmental variables were resampled to $256\cdot256$ (43 m/pix) to match the resolution of the satellite imagery. It is to note that climatic variables are known to a coarser spatial resolution than desired (1 km/pix). %is it really to note? It was downscaled linearly.

The area of Earth covered by the samples is 22.38 Million km$^2$. Our dataset accounts approximately for 10\% of the emerged surface of the Earth. Although good in size, the dataset is not perfect; the representation of biomes is unbalanced: due to the low cloud cover requisite deserts are overrepresented; some of the input environmental conditions are highly collinear; some samples cover mainly sea surface as some satellite intakes might be close to coastal areas; and there are a few areas of the planet where one or several predictors have missing values.  %Nonetheless, for a project of this size it is practially imposible to get the perfect dataset. In adition, we believe one of the requisite of the model should be to deal with such difficulties and learn the significant relations.

%%%%%%%%%%%%%%***Release the dataset: 166 Gb makes it difficult!***\\
%%%%%%%%%%%%%%***It is set and ready on the work2/SentMegadata folder, still not sure how the releasing can be done***

%[[LOTS OF PREPROCESSING IN USALLY HOWEVER...The high collinearity of climatic inputs was not dealt with, we use a complex deep learning model in order to avoid preprocessing biases. The network has to learn what features to extract from the inputs. In the same manner, there is an slight overrepresentation of deserts in the dataset:one of our assumptions is that the network will learn how to best assimilate the data.]]

\subsection{Models}\label{sec:models}
Two cGANs were trained, a low complexity one (fewer learnable filters per layer) and a high complexity one, named after the total size of the weights on disk. The small model has $57$ million weights (accounting for generator and discriminator) named `GAN 1 Gb' the high complexity model has $348$ million weights, the `7 Gb' model. 

The `1 Gb' GAN (low complexity) consists of a generator with symmetric encoder and decoder, skip connections (U-net) and a convolutional discriminator. It has 7 layers of two-dimensional convolutions with the following number of learnable $4×4$ filters: $64-128-256-512-512-512-512-512$ and exactly symmetrical deconvolutional decoder with $0.5$ dropout on the last three layers. Leaky relu activation and batch normalization is used at every convolutional step on the encoder. Relu activation and batch normalization at every deconvolutional step on the decoder except for the output layer that uses hyperbolic tangent activation. The discriminator consists of 5 convolutional layers with $64-128-256-512-1$ filters of size $4×4$, leaky relu activation and batch normalization at every step and sigmoid activation on the output layer. The `7 Gb' (high complexity) GAN model only differs on the number of filters on the generator ($160-320-640-1280-1280-1280-1280-1280$) and discriminator ($160-320-640-1280-1$).

Alongside with the cGANs, a fully connected model lacking spatial context was trained as a baseline. The fully connected network takes the value of the environmental variables as scalar inputs (32 input variables) and performs regression on the reflectance value for each spectral band (RGB and near infrared). The network has three fully connected hidden layers (64, 256, 364) using hyperbolic tangent activation for the hidden layers and a linear activation for the last layer. The fully connected network totals ~113,000 parameters. In addition, two handicapped cGAN models were trained in order to compare the cGANs to other models with the same complexity but lacking one of the key features. On one hand, a handicapped cGAN was trained over a modified train set with no spatial features due to random permutation of pixels. It is to expect that this handicapped cGAN will not take advantage of convolutions, since there are no spatial relations on the training set. On the other hand, the second handicapped cGAN was deprived of the discriminator loss, i.e., it was trained solely based on mean squared error. In this case we would expect it to fail to produce landscapes with sharp photointerpretable features.

%\section{Code}
%A python script that takes external arguments for training, generating and testing is attached. During execution it generates an \textit{.html} file that displays all predictors, generated and target samples for selected \textit{freq} as well as tensorboard logs. Evaluation software is not included as FRAGSTATS v4 \cite{mcgarigal2012fragstats} is a third party solution. Trained models (10 Gb) and full dataset as binary \textit{.tfrecords} can be provided under request (166 Gb).

\subsection{Supporting Figures}
% Include only the SI item label in the paragraph heading. Use the \nameref{label} command to cite SI s in the text.
% Place figure captions after the first paragraph in which they are cited.
Landscapes are segmented into landcover units before calculating patch level metrics. The following image depicts several landscape images together with their unsupervised segmentation (K=20)
\begin{figure}[H]
	\label{fig:segments}
	\centering
	\includegraphics[scale=1.3]{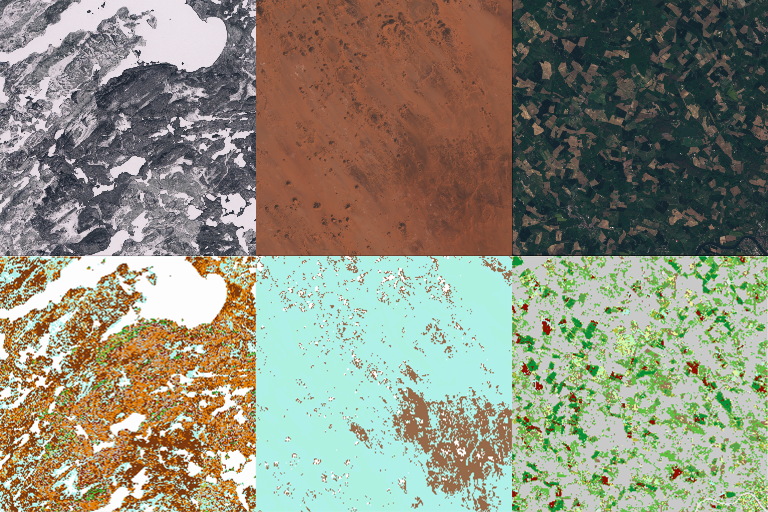} %Do not include figures on the submission TeX and PDF
	\caption{{\bf Landscape segmentation}
		Example landscapes and their unsupervised pixel-wise segmentation ($k=20$).}
\end{figure}

Further replicates of the main quantitative analysis for undersegmented secenario (K=8) and oversegmented scenario (K=60) are displayed next.
\begin{figure}[H]
	\includegraphics[scale=0.87]{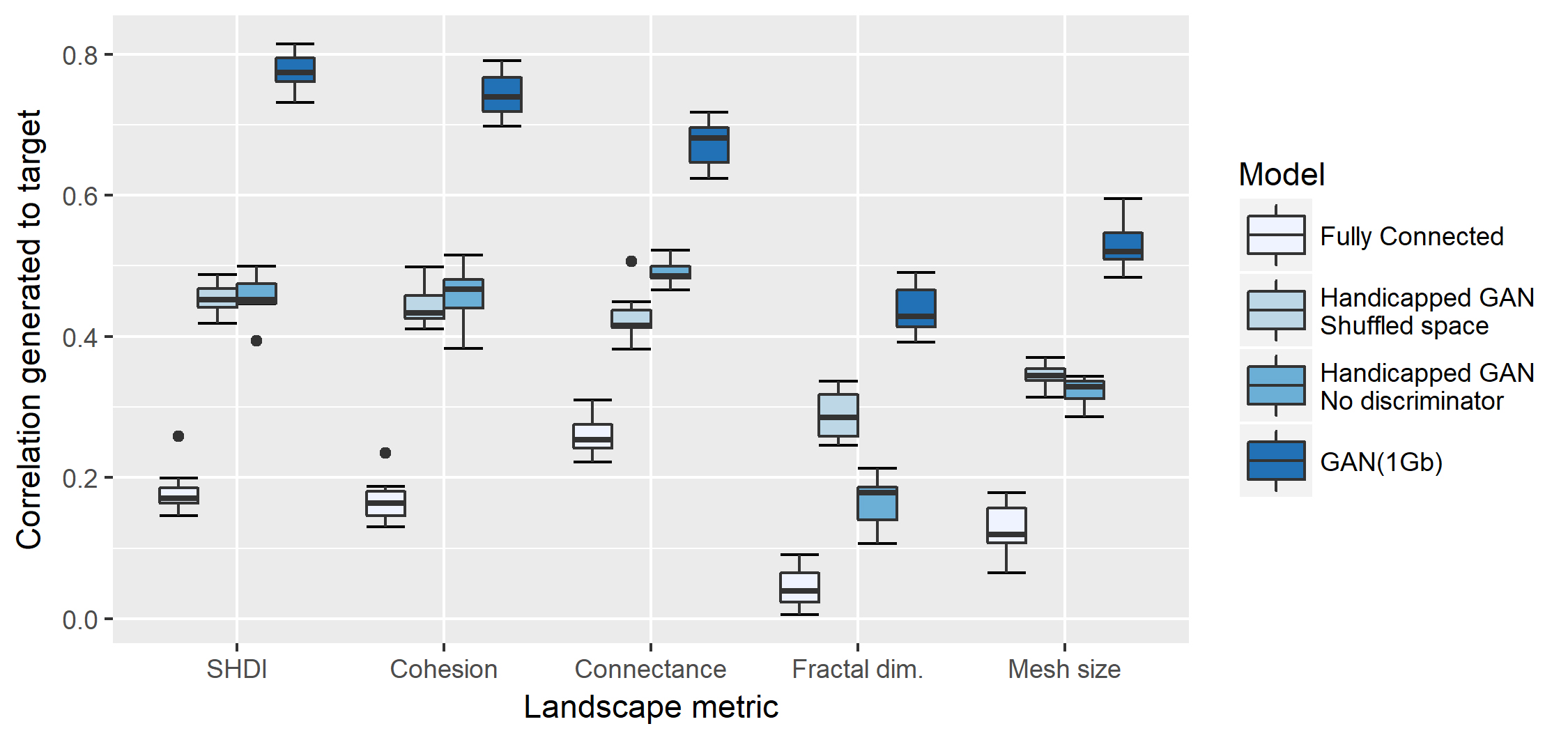} %Do not include figures on the submission TeX and PDF
	\caption{{\bf Intermodel comparison (8 landcover units)}
		Robust biweight midcorrelation between landscape metrics of generated and target landscapes. Shannon diversity index (SHDI), patch cohesion, patch connectance, average patch fractal dimensionality and effective mesh size were computed for both, real landscapes of test locations, and landscapes generated given the environmental conditions on the test locations. Landscapes were segmented into $8$ different patch types, serving as an undersegmentation case, i.e., the number of landcover units is excesivly low.}
	\label{fig:models_metrics_comparison_k8}
\end{figure}
\begin{figure}[H]
	\includegraphics[scale=0.87]{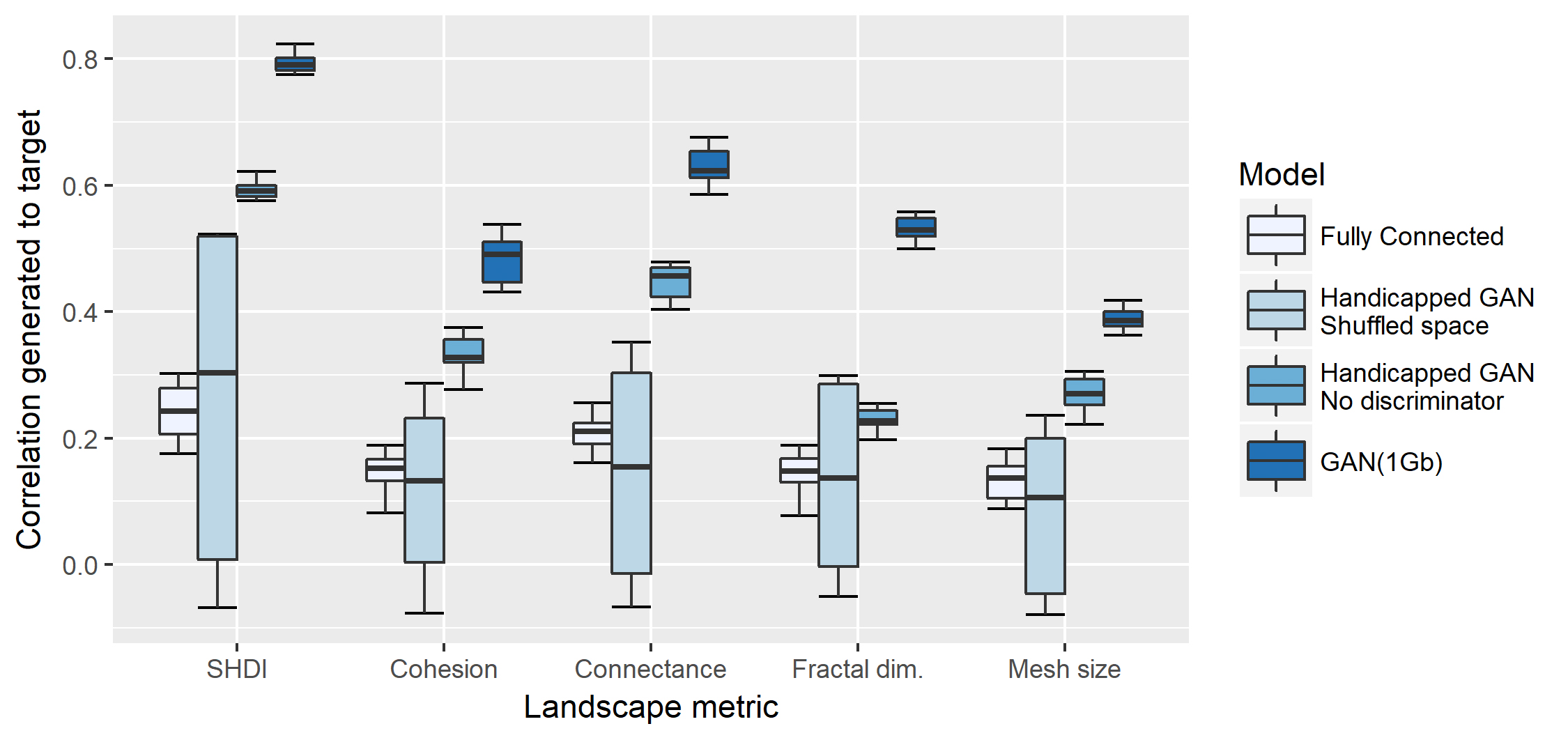} %Do not include figures on the submission TeX and PDF
	\caption{{\bf Intermodel comparison (60 landcover units)}
		Robust biweight midcorrelation between landscape metrics of generated and target landscapes. Shannon diversity index (SHDI), patch cohesion, patch connectance, average patch fractal dimensionality and effective mesh size were computed for both, real landscapes of test locations, and landscapes generated given the environmental conditions on the test locations. Landscapes were segmented into $60$ different patch types, serving as an oversegmentation case, i.e., the number of landcover units is excesivly high.}
	\label{fig:models_metrics_comparison_k60}
\end{figure}

\end{document}